\documentclass[letterpaper, 10 pt, conference]{ieeeconf}
\IEEEoverridecommandlockouts
\usepackage[utf8]{inputenc}
\usepackage{graphicx}
\usepackage{multicol}
\usepackage{multirow}
\usepackage{array}
\usepackage{footmisc}
\usepackage{amssymb}
\usepackage{amsmath}

\usepackage{amsthm}
\let\labelindent\relax
\usepackage{enumitem}
\usepackage{booktabs}
\usepackage{amsfonts}
\usepackage{algorithm}
\usepackage{algpseudocode}
\usepackage{tikz}
\usepackage{float}
\usetikzlibrary{math,shapes}
\usepackage{tabularx}
\usepackage{pbox}
\usepackage{ragged2e}
\usepackage{multirow}
\usepackage{tabularx}
\usepackage{balance}
\usepackage{subfig}
\usepackage{relsize}
\usepackage[colorlinks,
            linkcolor=blue,
            citecolor=MidnightBlue,
            urlcolor=blue]{hyperref}

\usepackage{listings}
\mathchardef\mhyphen="2D

\newtheorem{Pro}{Problem}

\newtheorem{Def}{Definition}

\newcommand{\SimName}{Botany-Bot}

\usepackage{xcolor}
\usepackage[normalem]{ulem} %

\usepackage{etoolbox}
\makeatletter
\patchcmd{\@makecaption}
  {\scshape}
  {}
  {}
  {}
\makeatletter
\patchcmd{\@makecaption}
  {\\}
  { -\ }
  {}
  {}
\makeatother
\usepackage{siunitx}
\title{\LARGE \bf 
\SimName{:} Digital Twin Monitoring \\
of Occluded and Underleaf Plant Structures with Gaussian Splats
}
\author{Simeon Adebola$^{1}$, Chung Min Kim$^{1}$,   Justin Kerr$^{1}$, Shuangyu Xie$^{1}$,\\ Prithvi Akella$^{2}$, Jose Luis Susa Rincon$^{2}$, Eugen Solowjow$^{2}$, and Ken Goldberg$^{1}$
\thanks{$^{1}$The AUTOLab at UC Berkeley (automation.berkeley.edu) {\tt\footnotesize \{simeon.adebola, goldberg\}@berkeley.edu}}%
\thanks{$^{2}$Siemens Research Lab, Berkeley, CA {\tt\footnotesize \{eugen.solowjow@siemens.com\}}}%
}

\begin{document}

\maketitle
\thispagestyle{empty}
\pagestyle{empty}

\begin{abstract}
Commercial plant phenotyping systems using fixed cameras
cannot perceive many plant details due to leaf occlusion. In this paper, we present \SimName{}, a system for building detailed ``annotated digital twins" of living plants using two stereo cameras, a digital turntable inside a lightbox, an industrial robot arm,  and 3D segmentated Gaussian Splat models. We also present robot algorithms for manipulating leaves to
take high-resolution indexable images of occluded details such as stem buds and the underside/topside of leaves. 
Results from experiments suggest that \SimName{} can segment leaves with 90.8\% accuracy, detect leaves with 86.2\% accuracy, lift/push leaves with 77.9\% accuracy, and take detailed overside/underside images with 77.3\% accuracy. Code, videos, and datasets are available at \href{https://berkeleyautomation.github.io/Botany-Bot/}{https://berkeleyautomation.github.io/Botany-Bot/}.

\end{abstract}

\section{Introduction}

Plant phenotyping platforms measure information about plants enabling practitioners to improve their understanding and make decisions that can improve yield. %
\cite{poorter2023pitfalls, roland_past_present}. 

Recently, emerging 3D reconstruction techniques building on Neural Radiance Fields~\cite{mildenhall2020nerf} and  Gaussian Splatting(GS) ~\cite{kerbl3Dgaussians} have demonstrated impressive visual reconstruction quality \cite{smitt2023pag, meyer2024fruitnerfunifiedneuralradiance, taim_ICRA}.

We propose \SimName{}, a system requiring only 4 cameras, a lightbox, a digital turntable, and a robot arm to obtain high-quality 3D digital twins of living plants. We use Gaussian Splatting to rapidly reconstruct 3D models which can be viewed in High Definition(HD) at a high frame rate of 30 frame per seconds, and extend our previous work on 
GARField~\cite{garfield2024} to segment GS models into leaves and stems. \SimName{}, includes an autonomous system with a 7-DOF robot arm to interact with individual leaves and take high-resolution images of their occluded surfaces (e.g. lift the individual leaf for leaf backside), capturing valuable additional detailed images for phenotyping, breeding, potential pest control, disease inspection, and growth monitoring \cite{MITSANIS2024108733,atefi_robotic_2021}.

We evaluate \SimName{} on a total of  109  leaves from  8  real plants. To assess performance, we define four key metrics: (1) the accuracy of automatically calculated physical metrics: height and leaf area, (2) the number of leaves successfully segmented and detected (zero-shot leaf detection and segmentation), (3) the number of leaves successfully manipulated, and (4) whether the leaf occluded side is fully visible after manipulation. 

\begin{figure}
    \centering
    \includegraphics[width=\linewidth]{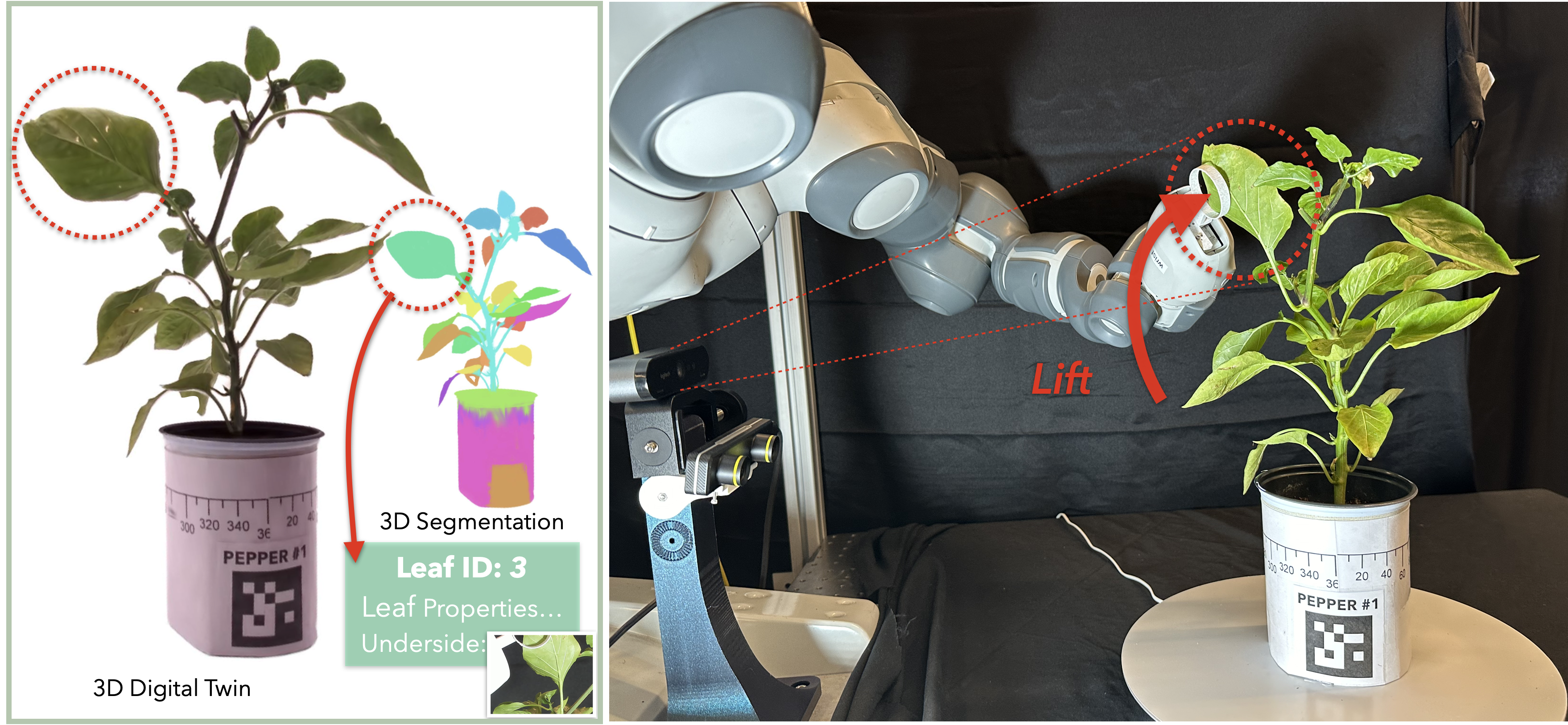}
    \caption{\textbf{Botany-Bot} creates detailed 3D digital twins of plants with a turntable and fixed cameras, segmenting them into individual components. \SimName{} then augments them by lifting or pushing down individual leaves to capture high-resolution images of the occluded sides of leaves. \SimName{} can also compute plant metrics; for example, the highlighted leaf has a leaf area of 25.6$\text{cm}^2$, leaf length of 7.2cm, and ground height of 27.2cm.}
     \vspace{-.15in}
     \label{fig:splash}
\end{figure}

\begin{figure*}[t]
    \centering
    \includegraphics[width=0.8\linewidth]{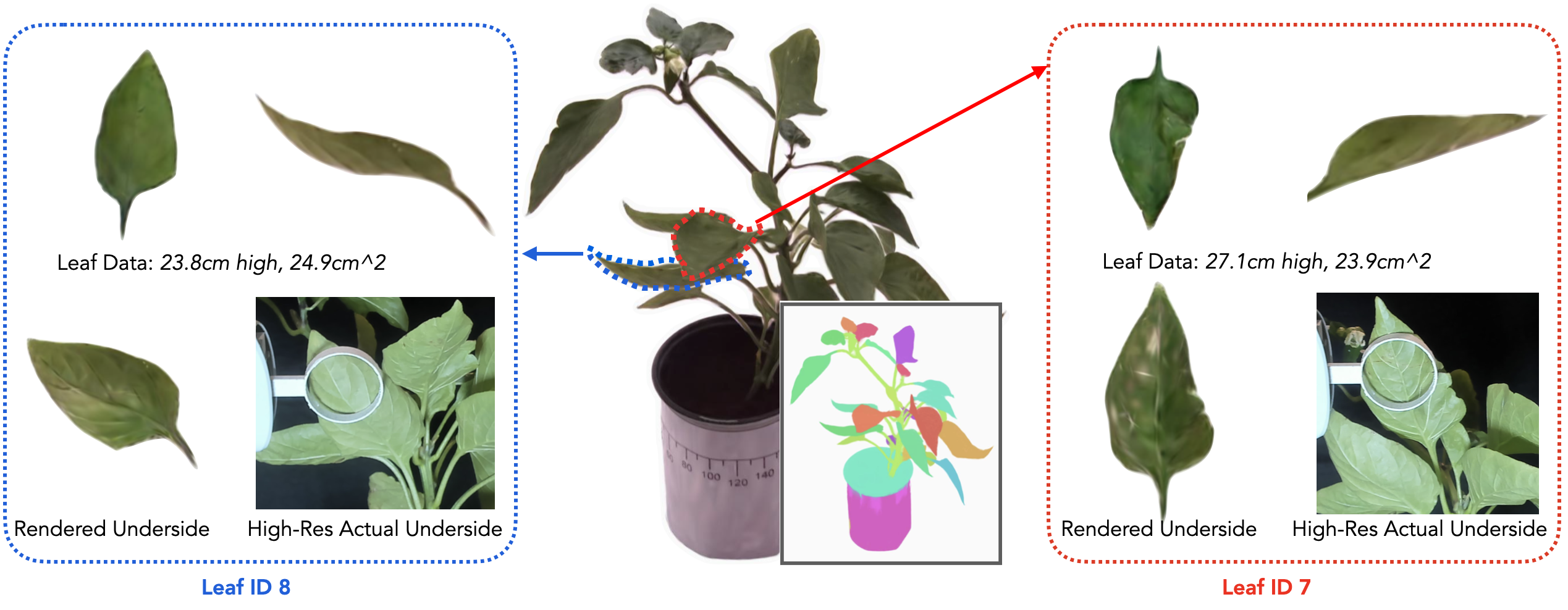}
    \caption{\textbf{Annotated 3D Digital Twins}: Botany-Bot produces high-fidelity 3D plant models along with segmented leaves (middle). By inspecting individual leaves, these models can be used to obtain physical properties like leaf height and area. In addition, the model is enhanced with robot interaction data where a robot arm lifts each leaf and examines its underside in high resolution. Pictured are examples of two leaves.}
    \label{fig:digital_twin_big}
\end{figure*}

This paper makes four contributions:
\begin{enumerate}
    \item \SimName{:} A novel pipeline for autonomously creating high-fidelity 3D digital twins of plants.
    \item A novel digital twin model for agricultural phenotyping that segments Gaussian Splats into constituent parts. 
    \item Algorithms for a robot to autonomously manipulate leaves to capture additional high-resolution images of their occluded surfaces.
   \item Results from experiments suggest that the pipeline is accurate in measuring the physical properties of the plant components.
   
\end{enumerate}

\section{Related Work}
\subsection{3D Reconstruction}
\label{subsection:recon_lit}
Neural Radiance Fields (NeRFs) \cite{mildenhall2020nerf} achieve ground-breaking 3D reconstruction quality of objects. NeRFs require images with known poses and creat 3D reconstruction by optimizing a volumetric representation~\cite{mildenhall2020nerf, nerfstudio}. 
NeRFs have been used in several fields including robot grasping \cite{pmlr-v164-ichnowski22a, pmlr-v205-kerr23a, lerftogo2023, shen2023F3RM}.
However, NeRFs are generally slow to train due to their use of an implicit neural network. Recently, \cite{kerbl3Dgaussians}~introduced three-dimensional (3D) Gaussian Splatting (3DGS) which achieved scene reconstruction and novel view synthesis comparable to NeRFs but with a much quicker rendering time. Gaussian splatting has found use in a similar range of fields as NeRFs including grasping \cite{zheng2024gaussiangrasper3dlanguagegaussian}.

\subsection{Plant Phenotyping }
\label{subsection:plantpheno}
For plant phenotyping, researchers acquire height, width, mass, size, and area for various plant parts, including fruits, leaves, and stems \cite{Esser2023field3D}. These information allows practitioners to make informed decisions for breeding to increase plant growth, recover from pests and diseases, and maintain a healthy plant ecosystem. Traditional methods for plant phenotyping, including segmentation, used 2D information collected through sensors such as cameras and LiDAR \cite{ag_tase}.
However, 2D phenotyping often has limited performance in tracking leaf-level details due to occlusion \cite{gibbs2018plant}.
As a result, 3D methods are increasingly favored for plant phenotyping for less occlusion \cite{paulus2019measuring, harandi2023make}. such as \cite{magistri2024icra} applying a transformer for 3D fruit shape completion, \cite{magistri2022ral-iros} using a neural network for fruit shape prediction, and \cite{pan2023iros} using a mobile robot for 3D mapping and joint prediction of fruit shape and pose.
\cite{Lehnert_3DMTS} moves a 3D camera array on a robot arm to determine the best place from which to take the next picture.

Other recent applications of 3D reconstruction techniques to agriculture include PAg-NeRF, which presents a NeRF-based pipeline for segmentation in 3D applied to agriculture \cite{smitt2023pag}. While PAg-NeRF \cite{smitt2023pag} shows good reconstruction quality, it focuses on fruit segmentation while \SimName{} uses Gaussian Splats and robot interaction to capture leaf structure. Esser et al. \cite{Esser2023field3D} use a mobile field robot outfitted with two laser scanners and a dome of 20 cameras to phenotype of plants in the field and use a neural implicit field method \cite{rosu2023permutosdffastmultiviewreconstruction} for 3D reconstruction. 
It does not segment the detected plants or reveal occluded areas.

\subsection{Robot-Plant Interaction and Motion Primitives}
Robot-plant interaction can enhance
a digital twin \cite{r2018research}. Design of robotic motions for plant interaction must account for the unique characteristics of plants, including their irregular shapes, deformable structures, and elastic properties. Existing research on plant manipulation focuses primarily on structurally destructive tasks, such as mechanical weeding \cite{mccool_robotic_weeding_tools,michaels_stamping}, fruit harvesting \cite{birrell2020field}, and sample collection \cite{mueller2017robotanist}. However, for non-destructive interactions, such as plant inspection, gentle and precise robotic movements are essential, particularly when interacting with delicate structures like leaves. Developing such inspection primitives remains an open challenge in robotic plant interaction.
\section{\SimName{} System Design}

\SimName{} introduces a low-cost plant scan system designed to capture and analyze potted plants using off the shelf hardware and software. The system uses static cameras and a digital turntable for multi-view imaging, enabling high-quality 3D reconstructions of plants without the need for extensive camera arrays. \SimName{} also incorporates robotic inspection that uses a detailed digital twin to autonomously examine plant features, such as leaf undersides, with precision and clarity beyond the initial reconstruction. 

\subsection{Hardware System}
\subsubsection{Plant Scan System}
\begin{figure}[t!]
    \vspace{0.2cm}
    \centering
    \includegraphics[width=\linewidth]{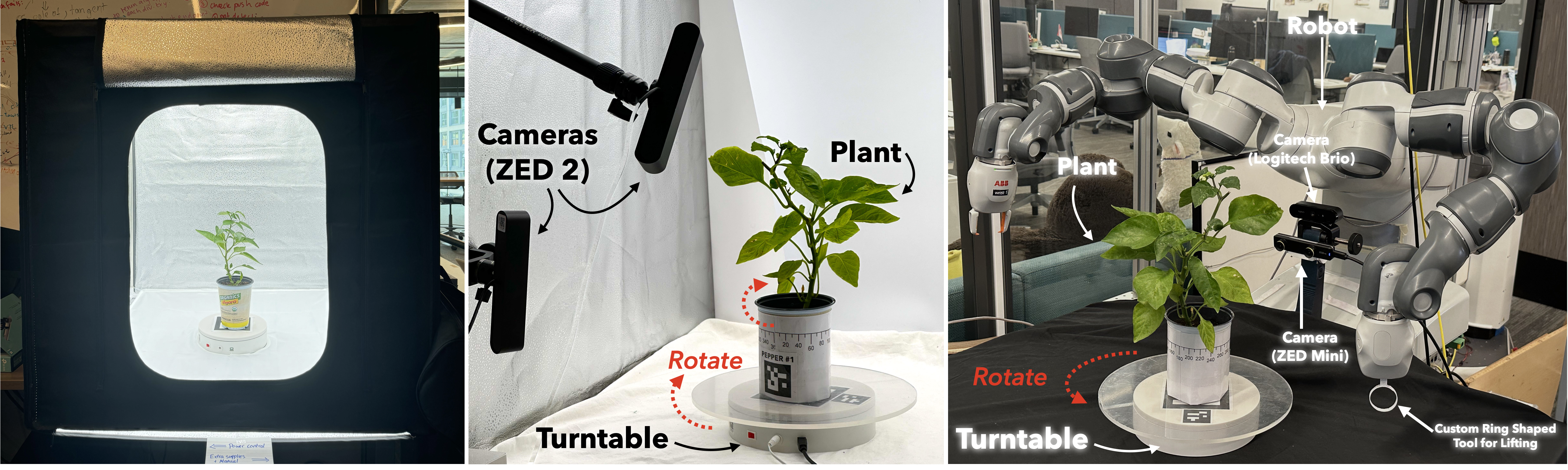}
    \caption{\textbf{\SimName{} Plant Scan, Inspection \& Interaction System.} Multi-view data collection is contained within a lightbox (left) which ensures uniform directional illumination for NeRF reconstruction. (Middle) a view inside the lightbox with turntable, plant, and 2 cameras. (right) the robot setup used for rotating the plant, lifting/pushing leaves, and imaging the undersides/tops of plant. The robot uses a digital turntable to minimize occlusion while manipulating the plant.}

    \vspace{-0.5em}
\label{fig:lightbox}
\end{figure}

\label{subsection:data_collection}
The plant scan system uses a commercially-available well-lit lightbox and two fixed cameras with a digital turntable to obtain a \textit{quasi-multiview} plant capture by rotating the potted plant around as shown in Figure~\ref{fig:lightbox}. This design allows 3D reconstruction techniques like NeRF\cite{mildenhall2020nerf} and 3DGS \cite{kerbl3Dgaussians} to obtain extensive multi-view coverage of the object being reconstructed. 

To obtain multi-view camera poses we place an ArUco marker~\cite{garrido2014automatic} on the turntable and calibrate the camera-to-turntable pose for angles in $0-360^{\circ}$ through which the turntable moves. Next, we place the plant on top of the turntable and repeat the same angles, which results in a multi-view posed capture. The digital turntable can be precisely controlled to turn by a specific angle, resulting in identical poses $\pm0.1^\circ$ to calibration. We utilize two ZED 2 Stereo cameras oriented vertically, for a total of 4 angles of elevation, and rotate the turntable to evenly spaced radial angles. The plant is positioned such that it is in view of all cameras, at a distance within the camera focal distance. Moreover, every plant also has an ArUco marker\cite{garrido2014automatic} which we use to save a relative pose between the plant and the turntable by calculating the relative pose between the camera-to-turntable pose and the camera-to-plant pose. 

\subsubsection{Plant Inspection \& Interaction System} \label{sec:inspection-sys}

The inspection system is composed of a high-resolution camera, another digital turntable to place the potted plant, and the robot arm to perform leaf manipulation equipped with a custom ring-shaped end effector to interact with the leaf while avoiding leaf obstruction. We refer to the custom ring-shaped end effector as the \textbf{inspection tool} in Sec. \ref{section:prob_statement} and Sec. \ref{section:method}. Since each plant is labeled with an ArUco tag, the relative pose from \ref{subsection:data_collection} allows registering the digital twin in the workspace of the robot. An example setup is shown in Figure~\ref{fig:lightbox}.

\subsection{Software System}
\begin{figure}[bthp!]
    \centering
    \includegraphics[width=0.8\linewidth]{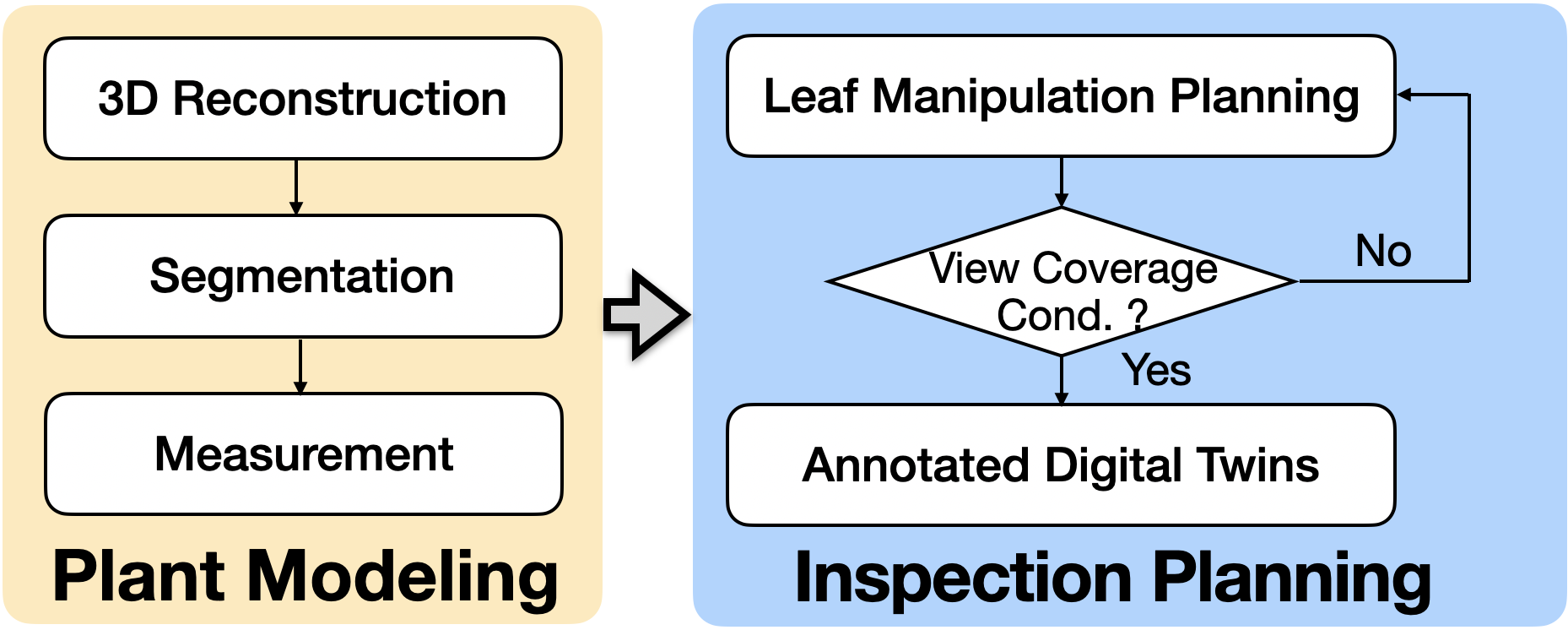}
    \caption{\textbf{Software Diagram for \SimName{}.} The modeling component takes plant images from multiple views, and models the plants by constructing in 3D, segmenting individual leaves and measuring the characteristics. Then the inspection component plans the leaf lifting, evaluates view coverage and we create annotated digital twins.}
    \label{fig:enter-label}
\end{figure}

Inside the lightbox, \SimName{} scans a potted plant using Sec.~\ref{subsection:data_collection} to capture multi-view images. From this sequence of images, it reconstructs a detailed Gaussian Splat model and segments it into its parts including leaves, stems, pot, flowers and fruits (where available). \SimName{} then uses this 3D model to plan leaf manipulation actions on a robot arm using the system in Sec.~\ref{sec:inspection-sys} to closely inspect finer details that were not captured in the 3D reconstruction. Finally, \SimName{} combines these images along with geometric statistics from 3D model analysis to create a data structure that includes all of these new views indexed by leaf ID. Figure ~\ref{fig:enter-label} shows the software diagram for the system.

\section{Problem Statement}
\label{section:prob_statement}
We assume that all plant leaves are separable, i.e. we do not consider dense grass or tightly packed leaves. 

\begin{Pro}[Digital Plant Initialization] \label{pro:plant_modeling}
    Given a plant, create $\mathcal{P}$.
\end{Pro}

\subsection{Output: Digital Twin Data Structure}
The data structure is defined by a set of four classes of components of the plant: leaf top, leaf bottom, side stem, main stem, as an indexing plant structure with the semantic and geometry information. We denote the plant component set as an ordered set $\mathcal{I} = \{1, ... ,N\},$ indexing by the component from the bottom of the main stem of the plant to the top of the plant where $N$ is the total number of components. $\mathcal{W}$ denotes the plant frame which is a right-handed 3D Euclidean coordinate system such that the coordinate origin aligns with the bottom center of the pot and the z-axis aligns with the plant growing direction (usually is the main stem direction).

For each component of the plant $ i\in \mathcal{I}$, we define a plant spatial feature set of the plant with $X_i = \{ \mathbf{x}_i , \mathbf{q}_i, c_i, \beta_i  \}$, where $\mathbf{x}_i = [x,y,z]^\textsf{T} \in \mathcal{W}$ is the center point of the component and $\mathbf{q}_i$ defines the direction of the component, such as the direction of the stem, the direction of the leaves of the principal axis. $c_i \subset \mathcal{C} $ is the class of component $i$, where $\mathcal{C} = \{\mbox{leaf top, leaf bottom, side stem, main stem}\}$. For each leaf, we also have $\beta_i \in \mathbb{R}^{n_s}$ describes $n_s$ shape parameters such as area size, leaf length. Then we define $\mathcal{P}$ as the digital twin:

\begin{equation}
    \mathcal{P} = \{X_i: i\in \mathcal{I}\} 
\end{equation}

\subsection{Plant Inspection}

To augment the digital twin 
with high-resolution images of occluded components, we define an inspection planning problem that computes a trajectory for the turntable and inspection tool to lift/push down leaves so that images taken from a front facing camera can observe it without occlusion.

\begin{Pro}[Inspection Planning] \label{pro:inspection}
Given a specified leaf ID $i \subset \mathcal{I}_{\mbox{\tiny leaf}} \subset \mathcal{I}$, find a leaf lifting / pushing motion for the inspection tool that enables unoccluded view.
\end{Pro}

\section{Methods}
\label{section:method}

\subsection{Creating the Digital Twin }
For Prob.~\ref{pro:plant_modeling}, we propose the algorithm framework as follows:

\subsubsection{3D Reconstruction}
\label{subsection:recon}
The rotating turntable multiview capture breaks the core fundamental assumption in NeRF and 3DGS that the scene remain static during capture in two ways: 1) the background around the object is static relative to the camera, and 2) lighting on the surface of the object is not 3D-consistent. To alleviate 1), we preprocess input data by automatically masking the potted plant with Segment Anything 2 (SAM 2)~\cite{ravi2024sam2segmentimages}. During radiance field construction, we do not compute standard loss functions on pixels lying outside this mask. We implement an extra L1 loss between the potted plant's mask and accumulation in the gaussian splatting reconstruction. Accumulation is a measure of the total accumulated alpha (opacity) at each pixel coming from multiple Gaussians. The lower the accumulation, the sparser the coverage with less Gaussians contributing to each pixel color. Therefore,  the L1 Loss is an important signal for the reconstruction to delete spurious geometry in the scene. We refer to this loss as an alpha loss.

To alleviate lighting inconsistency due to quasi-multiview data, we place the capture setup in a lightbox as shown in Figure \ref{fig:lightbox}, which results in highly directionally uniform lighting on the plant as it rotates. These modifications allows Botany-Bot to produce high quality plant reconstructions from quasi-multiview data, using 4 cameras to produce total views up to 4 times the number of evenly spaced radial angles turned to by the turntable. While we can take total views up to 4*number of turntable turns, we only save an image when there is a corresponding camera-to-turntable pose available as cameras may not always detect the aruco tag on the turntable for every turn during the initial pose calibration. So total views are sometimes less than 4*number of turntable turns. 

\subsubsection{Segmentation and Leaf Detection}
\label{subsection:seg_group}
\begin{figure*}[ht!]
    \centering
    \includegraphics[width=\textwidth]{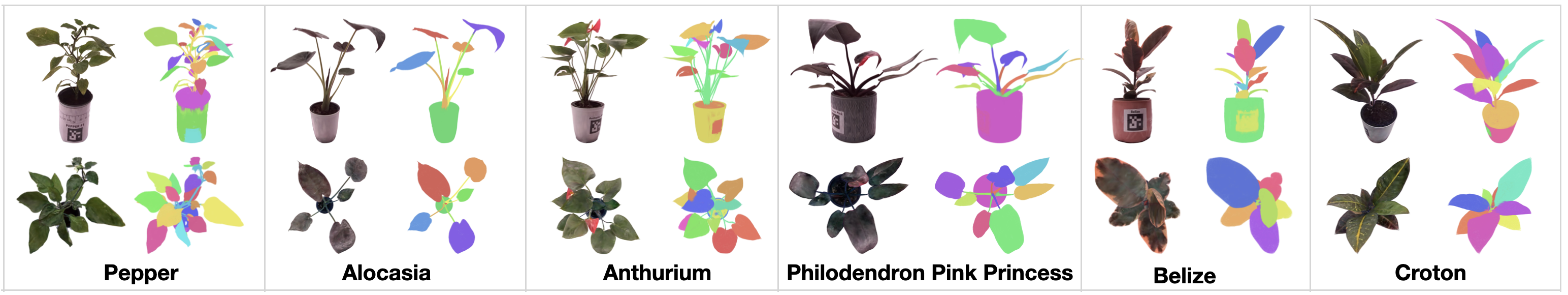}
    \vspace{-0.5cm}
    \caption[width=0.5\textwidth]{\textbf{\SimName{} Segmented Digital Twins:} Presented here are a side view and top view for six different plant species.  Each cell shows two different viewpoints of the 3D model, a side and top view. The RGB rendering is provided next to the colorized segmentation of plant components, each color corresponding to a different segment. Note both the visual fidelity of the models as well as the fine-grained leaf 3D segmentations.}
    \vspace{0.3cm}
\label{fig:nerfacto}
\end{figure*}

In addition to masking out the plant from the background to reconstruct it, we also wish to segment the plant apart into relevant sub-parts like leaves, stem. To accomplish this we use GARField~\cite{garfield2024}, a method for multi-level 3D segmentation which takes in SAM~\cite{kirillov2023segment} masks from multiple views and outputs a set of clusters corresponding to discrete groups. See Fig~\ref{fig:nerfacto} for example 3D segmentations (colored). \SimName{} automatically detects leaves by analyzing the 3D gaussians within each discrete segment: we filter for the pot by rejecting the three bottommost clusters (pot, pot texture, soil), two tallest clusters (pot, stem), and clusters composed of less than 100 points (noise). The remainder of the segments are deemed a leaf. Then, we estimate $X_i$ with its the center $\mathbf{x}_i$ across the 3D centers of Gaussians of the segments and its direction $\mathbf{q}_i$ using principal component analysis (PCA) \cite{abdi2010principal} to estimate the directions of the segmented point cloud. This approach is different from PAg-NeRF \cite{smitt2023pag} which provides a panoptic 3D representation which targets fruit segmentation.

\subsubsection{Plant Modeling and Analysis}
\label{subsection:plant_modeling}
From the digital twin we can extract physical properties of height, number of leaves, and total leaf area. Leaf height is calculated as the distance between the centroid of the leaf cluster and the bottom of the pot. The leaf surface area is calculated by leveraging the prior that leaves are large, round, and flat; we fit an oriented bounding box to the cluster, project along the smallest dimension, fit an ellipse to the 2D points, then calculate the area of the ellipse.
We estimate $\beta_i$ in $X_i$, i.e. leaf's length and width by the longest and second longest dimensions of the bounding box.

\subsection{Robot Primitives for Interaction and Leaf Lifting/Pushing}
\label{subsection:robot_inter}
Although the rotating capture procedure (Sec.\ref{subsection:data_collection}) collects multi-view images of the target plant, it cannot observe key areas such as leaf undersides, leading to under-reconstruction for the digital twin (Fig. \ref{fig:digital_twin_big} shows that rendered underside could be blurry). These regions are crucial for detecting issues such as pest infestations or diseases. To address this, \SimName{} uses robot interaction with a custom end-effector to lift/push down each leaf toward a static camera, capturing high-resolution underside/overside images (Fig.~\ref{fig:undersides}). Guided by the plant component data structure, we next address Problem~\ref{pro:inspection}.

\subsubsection{Task Primitives}
\label{subsubsection:task_primi}
There are three sequential steps for manipulating a target leaf using the robot arm with its inspection tool and turntable.
\begin{enumerate}[leftmargin=*]
    \item[a] \textbf{Rotation Alignment}: The target leaf is rotated through the rotation of the turntable by $\theta$ such that its principal axis (i.e. the stem direction for the leaf) $q_i$ is aligned to the high resolution camera z-axis within a small margin $\epsilon$ and the leaf center is on the camera z-axis. We denote the turntable rotation set that satisfies the alignment condition to be $\mathcal{A}_{\mbox{\tiny rotate}}\subset SO(2)$
\item[b] \textbf{Tool Positioning}: The inspection tool is moved directly above or underneath the leaf center to position it for lifting/pushing. This step ensures that the tool’s position  $p$  aligns with the leaf center in the horizontal plane: $\mathbf{x}_{\mbox{\tiny prepare}} = (x_i, y_i, z_{t_{\text{initial}}})$ where $z_{t_{\text{initial}}}$  is the initial height of the tool. The tool also needs to be parallel to the the leaf surface. 
 We denote the inspection tool pose set that the tool positioning satisfies $\mathcal{A}_{\mbox{\tiny prepare}}\subset SE(3)$. 

\item[c] \textbf{Manipulation}: The inspection tool moves upward/downward while simultaneously rotating to lift/push down the leaf. The upward motion follows: $\mathbf{x}_{\mbox{\tiny lift}} = (x_{i}, y_{i}, z_{t_{\text{final}}})$ where $ z_{t_{\text{final}}} > z_{t_{\text{initial}}}$ while the downward motion follows: $\mathbf{x}_{\mbox{\tiny push}} = (x_{i}, y_{i}, z_{t_{\text{final}}})$ where $ z_{t_{\text{final}}} < z_{t_{\text{initial}}}$. Simultaneously, the inspection tool rotates by an angle  $\phi $, applied as: $R_{\mbox{\tiny lift/push}} \subset SO(3)$ so that the leaf is lifted/pushed down in a controlled manner. The inspection tool pose set $\mathcal{A}_{\mbox{\tiny lift/push}} \subset SE(3)$ satisfies the lifting/pushing condition. 
\end{enumerate}

This structured sequence ensures a stable manipulation of the target leaf and we denote a feasible action set to be $\Pi = \{\pi = (a_1,a_2,a_3): a_1\in \mathcal{A}_{\mbox{\tiny rotate}}, a_2 \in \mathcal{A}_{\mbox{\tiny prepare}}, a_3\in \mathcal{A}_{\mbox{\tiny lift}} \}$. 

\subsubsection{Inspection Planning}\label{ssc:task-constraints}

In order to obtain a high quality image observation, we propose the leaf lifting/pushing trajectory generation for inspection as an optimization problem which maximizes view coverage of the plant sub-component.
\begin{Def}[View Coverage]
    For leaf $i$, denote the estimated visible leaf surface area in the camera field of view to be $\mbox{Area}(v_i, X_i)$ and the actual surface area to be $\mbox{Area}(s_i)$ of the plant segments $s_i$. View coverage is defined as: 
\begin{equation}
\label{eq:view_coverage}  
C(\pi,X_i) = \frac{\mbox{Area}(v_i, X_i)}{\mbox{Area}(s_i)}
\end{equation}
\end{Def}

The optimization function is formulated as:
\begin{align}
    \max_{\pi \in \Pi} ~ C(\pi,X_i)
    \label{eq:opt_fxn} 
\end{align}

\subsubsection{Annotated Plant Digital Twins}
\label{subsection:augment_views}
\SimName{} accumulates all of its physical property analyses, leaf detections, and visual observations into a single indexable model, whereby a user can view the plant in 3D, select any leaf, and view captured lifted images and physical data. An example visualization of this data structure is provided in Fig~\ref{fig:splash}.

\begin{figure}[t]
    \centering
    \includegraphics[width=0.5\linewidth]{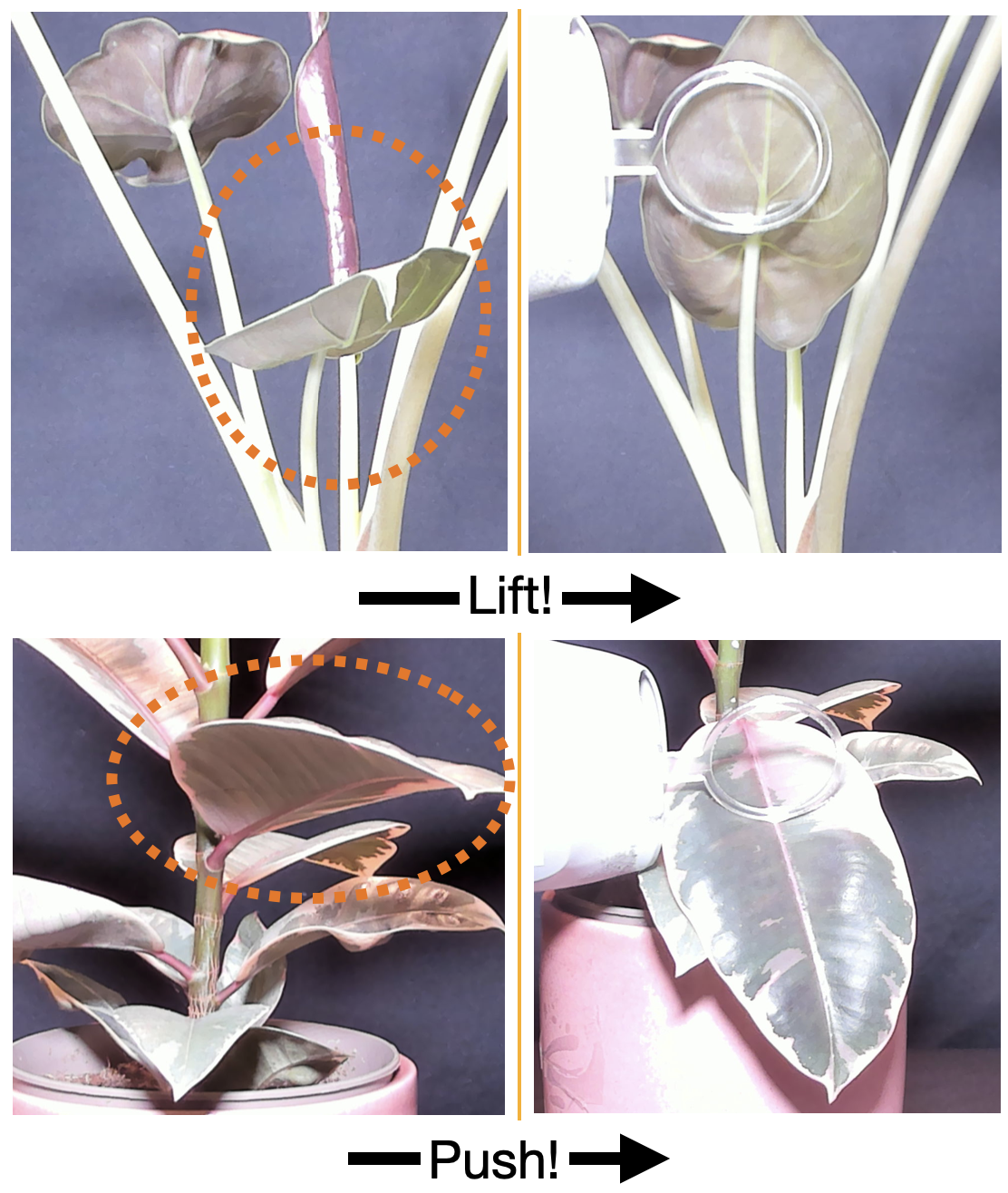}
    \caption{\textbf{Leaf Surfaces} revealed by robot manipulation for alocasia and pp-philodendron. Each left/right pair shows a leaf before and after lifting/pushing it. Previously self-occluded regions become substantially more visible after interactive inspection.} 
    \label{fig:undersides}
    \vspace{-1em}
\end{figure}

\section{Experiments}
We evaluate the following three metrics for 3D reconstruction:
\begin{enumerate}
    \item Ratio of leaves correctly \textit{segmented} on each plant
    \item Ratio of leaves successfully \textit{detected} per plant
    \item Physical accuracy of leaf area and leaf height on a select subset of leaves measured against ground-truth
\end{enumerate}
and these two metrics for autonomous robot leaf inspection:
\begin{enumerate}
    \item Number of leaves autonomously \textit{lifted} by robot
   \item Undersides of leaves fully \textit{revealed} to high-res camera.
\end{enumerate}

\subsection{Automatic Collection of Plant Properties}
\subsubsection{Experiment Setup}
For experiments, we use (a) three pepper plants (all of the California Wonder specie) (b)  Pink Princess Philodendron (PP-Philodendron) (c) Anthurium (d) Chroma Belize (e) Croton (f) Alocasia. All plants were procured from a local Home Depot store. These plants are chosen in line with our assumption that all plant leaves are not too densely cluttered and also that they are small enough to fit into the two experimental areas: lightbox(having been placed on the digital turntable), and robot+cameras'(where possible) workspace. 

For each plant, we collect a dense capture of the plant using Sec.~\ref{subsection:data_collection}. We set the digital turntable to an angle turn of 15\textdegree{} resulting in a total of 24 images (i.e. $360$\textdegree  / $15$\textdegree ) per camera for every complete revolution of the digital turntable and up to 96 total views. The value of 15\textdegree{} was empirically determined to allow sufficient view coverage of the plant on the turntable.

\begin{figure}[t]
    \centering
    \includegraphics[width=0.8\linewidth]{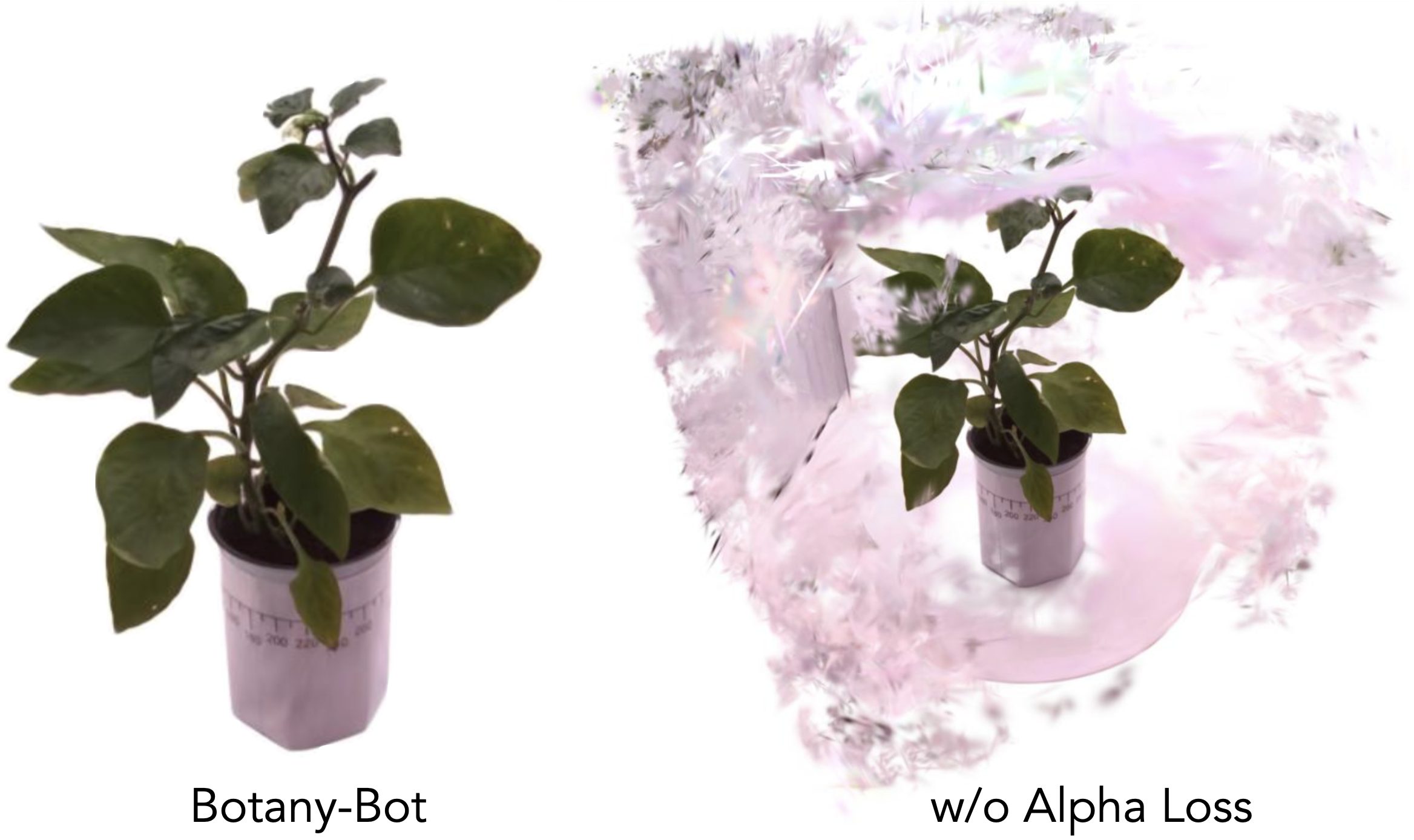}
    \caption{\textbf{Alpha loss ablation}: without implementing our alpha loss penalizing out-of-mask pixels (Sec.~\ref{subsection:recon}), the reconstruction drastically degrades in quality because the quasi multi-view introduces floaters.}
    \vspace{-.15in}
    \label{fig:alpha_loss}
\end{figure}

\begin{table}[t]
\footnotesize
\centering
\begin{tabularx}{\linewidth}{
  >{\raggedright\arraybackslash}l  %
  >{\raggedleft\arraybackslash}X   %
  >{\raggedleft\arraybackslash}X   %
  >{\raggedleft\arraybackslash}X   %
}
\toprule
\multirow{2}{*}{\textbf{Plant}} & 
\multicolumn{1}{l}{\textbf{Leaf Height}} & 
\multicolumn{1}{l}{\textbf{Leaf Area}} & 
\multicolumn{1}{l}{\textbf{Num. leaves}} \\
& \multicolumn{1}{l}{\textbf{(cm)}} &
\multicolumn{1}{l}{\textbf{(cm$^2$)}} &
\multicolumn{1}{l}{\textbf{}} \\
\midrule
Pepper \#1 & $25.4\pm6.4 $ & $18.4\pm15.3 $ & 22\\ 
Pepper \#2 & $25.8\pm5.7 $ & $16.1\pm7.8 $ & 13\\ 
Pepper \#3 & $31.5\pm4.5 $ & $11.2\pm8.5 $ & 21\\ 
PP-Philodendron & $23.3\pm3.2 $ & $46.1\pm13.5 $ & 6\\ 
Alocasia & $30.8\pm6.9 $ & $55.8\pm35.5 $ & 5\\ 
Anthurium & $27.3\pm6.8 $ & $38.2\pm12.2 $ & 11\\
Chroma Belize & $20.7\pm5.0 $ & $40.1\pm21.7 $ & 9\\ 
Croton & $19.7\pm4.4 $ & $27.8\pm10.5 $ & 7\\ 
\bottomrule
\end{tabularx}
    \caption{\textbf{Examples of physical metrics} that can be calculated from digital twin plant data, by calculating them directly over each 3D leaf segment as described in Sec.~\ref{subsection:seg_group}.}
\vspace{-1em}
\label{tab:leaf_metrics}
\end{table}

\subsubsection{Results}
Fig.~\ref{fig:nerfacto} shows the digital twins generated for six of the eight plants, using Splatfacto~\cite{nerfstudio,ye2023mathematical} for the 3DGS reconstruction and GARField~\cite{garfield2024} to group the gaussians on the plant into multiple segments. Each leaf can also be visualized separately and inspected in detail, with examples in Fig.~\ref{fig:digital_twin_big}. Fig. ~\ref{fig:alpha_loss} shows the effect of the added alpha loss penalizing out-of-mask pixels (Sec. IV-B), as without the loss,  reconstruction drastically degrades in quality because the quasi multi-view introduces floaters.
We study if the digital twin model provides sufficient information to reliably collect leaf statistics automatically. The plant segmentation and leaf detection results are shown in the first two columns in Table~\ref{tab:leaf_data_eval} (``Leaves Segmented" and ``Leaves Detected"). The method correctly segments 91\% and detects 91\% of the leaves, even without using a task-specific leaf segmentation model or navigating the multiple mask ambiguity in Segment Anything Model~\cite{kirillov2023segment}. The only leaves that remain unsegmented are very small in size ($\leq1\text{cm}$ in length), or very close to a dense area in the plant (e.g., where the plant branches into multiple stems, or the pot and soil). In a case where two leaves are segmented as one(e.g. Anthurium), we consider this a failure and do not count either leaf. 

Once the leaves are identified, the digital twin can be used to track the number of leaves, and their height and areas as shown in Table~\ref{tab:leaf_metrics}. To evaluate the accuracy of these measurements, we choose three random leaves per plant and manually measure their length and width using a ruler. The results are in Table~\ref{tab:leaf_data_eval}, with an average absolute error~(MAE) of $2.0\text{cm}$. The calculated values tend to underestimate the dimensions of the leaf, for two reasons: one, the leaf length for pepper includes parts of the connection to the stem, while GARField~\cite{garfield2024} tends to exclude this from the leaf segmentation; and two, the oriented bounding box-based method in Sec.~\ref{subsection:seg_group} does not account for any curvature in the leaf or any geodesic distances along the extents. 

\begin{table}
\centering
\begin{tabularx}{\linewidth}{
  l
  >{\raggedleft\arraybackslash}X
  >{\raggedleft\arraybackslash}X
  >{\raggedleft\arraybackslash}X
  >{\raggedleft\arraybackslash}X
}
\toprule

\multirow{2}{*}{\textbf{Plant}} &
\multicolumn{1}{l}{\textbf{Leaves}} &
\multicolumn{1}{l}{\textbf{Leaves}} &
\multicolumn{2}{l}{\textbf{MAE/MRE} (cm / \%)} \\
& \multicolumn{1}{l}{\textbf{Segmented}} &
\multicolumn{1}{l}{\textbf{Detected}} &
\multicolumn{1}{l}{\textbf{Length}} &
\multicolumn{1}{l}{\textbf{Width}} \\

\midrule
Pepper \#1 & 22/23 & 22/23 & 2.3/23.7   & 1.7/28.4 \\
Pepper \#2 & 13/17 & 13/17 & 2.4/23.5   & 1.2/25.6\\
Pepper \#3 & 21/22 & 21/22 & 2.2/17.0   & 2.1/34.4\\
PP-Philodendron & 7/7 & 6/7 & 2.4/17.5   & 0.8/12.4 \\
Alocasia & 6/6 & 5/6 & 3.5/24.1   & 1.4/17.6\\
Anthurium & 10/12& 11/12 & 3.0/23.9   & 0.9/11.3\\
Chroma Belize & 11/13 & 9/13 & 2.2/17.5   & 1.0/13.0 \\
Croton & 9/9 & 7/9 & 4.0/30.3   & 1.6/26.9 \\
\bottomrule

\end{tabularx}
\caption{\textbf{Evaluating Automatic Leaf Detection, Segmentation and Measurement:} We detect and segment leaves as described in Sec.~\ref{subsection:plant_modeling}. Next, for three random leaves for each plant, we estimate their length and width as described in Sec.~\ref{subsection:seg_group} (the longest and second longest lengths of the oriented bounding box extents). Note that ``Leaves Segmented" only looks at the segmentation, while ``Leaves Detected" looks at the true number of clusters decided to be leaves. We measure their ground-truth values using the real plants, and report the mean absolute and relative errors.}
\label{tab:leaf_data_eval}
\vspace{-2em}
\end{table}

\subsection{Leaf Lifting/Pushing}
\subsubsection{Experimental Setup}
We use the YuMi IRB 14000 robot arm \cite{ABB_yumi} for navigating to and lifting/pushing the leaves, and a logitech Brio camera \cite{logitech-brio} for taking images of the underside/overside. Depth from a Zed mini camera\cite{stereolabsMiniStereo} is used as a guide to manually align the plant pose in the world frame. For the lifting primitive, this is done once at the beginning for the sequence of manipulated leaves; while for the pushing primitive, alignment is done for every manipulated leaf. 
The lifting/pushing motion follows the procedure in Sec.~\ref{subsection:robot_inter}. The end effector approaches the leaf by targeting the centroid of its cluster, then lifts/pushes each leaf by 65\% to 90\% (empirically determined) of the longest dimension of its oriented bounding box, proportional to its size. Combining the task primitives in Sec. ~\ref{subsubsection:task_primi} comprises our heuristic solution to the optimization function in Equation \eqref{eq:opt_fxn}.
Robot motion planning is done using the Jacobi Motion Library\cite{jacobi}.

To evaluate \SimName{}’s interactive inspection capability, we measure the number of leaves lifted/pushed autonomously and the number of undersides/oversides fully revealed to the high-res camera. The first metric evaluates ability to interact without causing damage, while the second gauges motion accuracy. We define a leaf as successfully lifted if the leaf is lifted upwards/pushed downwards by the robot motion, and its underside/overside as fully observed if at least 75\% of the bottom/top surface is fully visible to the camera after interaction. At 75\% a significant majority of the leaf surface is fully visible. For leaves that are facing upwards and have the underside visible prior to the lifting motion, we count the motions and images as a success only if the observation has been improved by the interaction. We only lift leaves that are over 5cm in length, and try to filter any robot actions that may cause damage to the plant. The lift/push distance and the angle by which the inspection tool rotates are parameters determined by each plant. 

\subsubsection{Results}
Robot interaction results are reported in Table~\ref{tab:robot_metrics}; across 68 safely accessible leaves across 8 plants, the robot can successfully lift/push 53 leaves. Where the leaf is not sufficiently aligned with the camera, we mark this as a failure even if the lift/push motion is executed correctly. In one notable case (Croton), the robot successfully pushes down a leaf but the leaf is broken. We do not include this leaf among the considered results. Besides breakage, for lifting/pushing, there are three main failure cases: i. leaf obstructions, ii. plant dynamics, and iii. pose error. A robot may fail to interact with a leaf properly if the gripper accidentally catches on a lower leaf during its motion, bends the stem, and causes the leaf to rotate out of the way. Also, even if the robot made proper contact with the leaf initially, the leaf may slip out of the way depending on how it is connected to the stem. Any pose registration error will only exacerbate these problems. Solving this would require some form of visual closed-loop servoing to detect any errors.

Out of the 53 leaves that were pushed/lifted, their oversides/undersides are visible in 41 cases. Example images are presented in Figs.~\ref{fig:digital_twin_big} and \ref{fig:undersides}. The images are significantly more visually detailed, featuring the leaf veins as well as the node connecting the leaf to the rest of the stem. In contrast, the Gaussian Splat reconstructions are under-reconstructed at the overside/underside, with blotchy color artifacts. 

\begin{table}
\centering
\begin{tabularx}{\linewidth}{l >{\raggedleft\arraybackslash}X >{\raggedleft\arraybackslash}X >{\raggedleft\arraybackslash}X} %
\toprule
\textbf{Plant} & \multicolumn{1}{l}{\textbf{Lifted}} & \multicolumn{1}{l}{\textbf{Pushed}} & \multicolumn{1}{l}{\textbf{Observed}} \\ 
\midrule
Pepper \#1 & 9/15 & - & 9/9 \\ 
Pepper \#2 & 8/9 & - & 5/8 \\ 
Pepper \#3 & 8/10 & - & 2/8  \\
PP-Philodendron &  6/6 & - & 5/6  \\
Alocasia &  3/4 & - & 3/3 \\
Anthurium & 8/9 & - & 8/8 \\
Chroma Belize* & - & 8/9 & 6/8 \\
Croton* & - & 3/6 & 3/3\\
\bottomrule
\end{tabularx}

 \caption{\textbf{Evaluating \SimName{} Robot-Leaf Interactions}. We only lift/push down (*indicate push motion) leaves we find are possible by the motion planner without damaging the robot, and those that are visible in the camera already. Note that although a leaf may be lifted or moved, this does not guarantee that the leaf underside is visible to the camera.}
\label{tab:robot_metrics}
\vspace{-3em}
\end{table}

For observing leaf oversides/undersides, the biggest challenges are singulating the leaf and choosing the correct distance to lift/push down the leaf. Most failure cases (6/12) are due to a nearby leaf that gets lifted up/pushed down, and blocks the underside of the target leaf from the camera view. Collision-free, contact-aware motion planning with the 3D plant model would be required to ``burrow" between leaves carefully. Another failure mode (5/12) is the leaf being lifted/pushed but not enough to expose its underside/overside. This is because the leaf lift/push distance selection is a naive implementation; the lift/push distance should depend on the leaf's position in the camera image center, as the leaves lower on the camera perspective should be lifted up a larger amount while leaves higher on the camera perspective should be pushed up a larger amount. Another solution could be to implement a closed-loop motion that takes a image just before losing leaf contact. Lastly, we notice that the gripper did not always orient parallel to the leaf surface. Improving the plant-specific tuning of this parameter could help fix this. 

\subsection{Cost Analysis}
The lightbox, 2 Zed 2 stereo cameras and digital turntable have a total cost of \$1,137.88 (1,025.46 Euros). Apart from the robot, the two cameras cost \$598.99 (539.81 Euros) bringing the total cost of the system apart from the robot to \$1,736.87 (1,565.27 Euros). The YuMi Bimanual robot\cite{ABB_yumi} costs $\$71,277$ for a full configuration, according to an email received from ABB on March 5, 2025. In this instantiation of \SimName{}, only one arm of the bimanual robot is used.

\section{Limitations and Future Work}
\SimName{} is not designed for tightly clustered or small leaves. Incorporating more precise motion planning or closed-loop servoing for interactive leaf lifting is a subject for future work. Segmentation accuracy may be improved by tuning SAM 2 with plant data or by exploring alternate approaches like learning ultrametric feature fields for hierarchical clustering in 3D segmentation\cite{haodi_ultrametric}. The 3D reconstructed digital plant could also potentially be used to test motion planning and pruning to avoid damaging the real plant. We also note that currently we do not update the 3D models of the actual reconstructions with the new high definition images taken after robot interaction but only annotate the models using the images. Updating the actual reconstructions is still future work. Additionally, we only use one arm of the bimanual robot arm with a custom inspection tool, and in future work, we will extend to bimanual manipulation. We will also consider a new generation of low-cost robot platforms like the ALOHA\cite{Zhao-RSS-23, zhao2024alohaunleashedsimplerecipe} or YAM\cite{yam_robot} platforms. Moreover, a user study of \SimName{} will also be useful to help evaluate how useful agricultural practitioners find the system for their phenotyping needs.
Finally, future work will explore capturing and maintaining agricultural digital twins over time to study growth, potentially taking advantage of dynamic multi-view reconstruction methods. 
\section*{Acknowledgements}
\vspace{-0.1cm}
\relsize{-1}
This research was performed at the AUTOLAB at UC Berkeley in affiliation with the Berkeley AI Research (BAIR) Lab. This research was supported in part by Siemens. Simeon is supported in part by the Bakar BioEnginuity Impact Grant. We thank our colleagues who provided helpful feedback and suggestions, in particular Shrey Aeron, Brent Yi, Justin Yu, and Kush Hari.

\bibliographystyle{IEEEtran}
\bibliography{IEEEabrv,references}

\end{document}